\documentclass[runningheads]{llncs}
\usepackage[T1]{fontenc}
\usepackage{graphicx}
\usepackage{xcolor}
\usepackage{algorithm}
\usepackage{algorithmicx}
\usepackage[noend]{algpseudocode}
\usepackage{amsfonts}

\begin{document}

\title{A Study of Data-driven Methods for Adaptive Forecasting of COVID-19 Cases}

\author{Charithea Stylianides\inst{}\orcidID{0009-0002-3568-3449} \and\\
Kleanthis Malialis\inst{}\orcidID{0000-0003-3432-7434} \and\\
Panayiotis Kolios\inst{}\orcidID{0000-0003-3981-993X}\thanks{This work was supported by the European Union's Horizon 2020 research and innovation programme under grant agreement No 739551 (KIOS CoE - TEAMING) and from the Republic of Cyprus through the Deputy Ministry of Research, Innovation and Digital Policy. It was also supported by the CIPHIS (Cyprus Innovative Public Health ICT System) project of the NextGenerationEU programme under the Republic of Cyprus Recovery and Resilience Plan under grant agreement C1.1l2.}}

\authorrunning{C. Stylianides et al.}
%
\institute{KIOS Research and Innovation Center of Excellence, University of Cyprus, Nicosia, Cyprus 
\email{\{stylianides.charithea, malialis.kleanthis, kolios.panayiotis\}@ucy.com.cy}}

\maketitle             

\begin{abstract}
Severe acute respiratory disease SARS-CoV-2 has had a profound impact on public health systems and healthcare emergency response especially with respect to making decisions on the most effective measures to be taken at any given time. As demonstrated throughout the last three years with COVID-19, the prediction of the number of positive cases can be an effective way to facilitate decision-making. However, the limited availability of data and the highly dynamic and uncertain nature of the virus transmissibility makes this task very challenging. Aiming at investigating these challenges and in order to address this problem, this work studies data-driven (learning, statistical) methods for incrementally training models to adapt to these nonstationary conditions. An extensive empirical study is conducted to examine various characteristics, such as, performance analysis on a per virus wave basis, feature extraction, ``lookback'' window size, memory size, all for next-, 7-, and 14-day forecasting tasks. We demonstrate that the incremental learning framework can successfully address the aforementioned challenges and perform well during outbreaks, providing accurate predictions.

\end{abstract}

\section{Introduction}
The COVID-19 pandemic has caused a massive disruption to society since its emergence in December 2019. An unprecedented number of people were infected, hospitalized and had COVID-19 being their leading cause of death. Moreover, the consequences of the pandemic are still impacting our social and economic ecosystems. Evidently, many countries still impose restrictions and measures based on the evolution of the infected population. Hence, effective modelling and prediction of the evolution of the viral load in the society can be of detrimental factor in decision making. By taking proactive measures for closures and lockdowns, restricting public events, health guidelines and vaccine policies, governments can increase their effectiveness and limit transmissibility. A way to capture the spread of the virus is by tracking and predicting the number of positive cases. This constitutes a challenging task because of:

\textbf{Data non-stationarity}. The data exhibit a highly dynamic behaviour, i.e., the data distribution evolves over time \cite{ditzler2015learning}. In the COVID-19 case, for instance, there have been many variants of the virus (e.g., Delta and Omicron), as well as many measures which have been imposed (e.g., vaccination and school closure).

\textbf{Limited data}. This refers to the problem of having limited availability of historical data. Evidently most countries reported positive cases on a daily basis which accumulates to a mere 365 data points over the course of a year.

As a result, it is necessary to have an online learning model which is able to adapt to non-stationary environments, and to be incrementally trained from limited data. The contributions of this work are the following.
\begin{itemize}
    \item The primary focus of this study is on COVID-19 cases forecasting for Cyprus, a European country with a population of around one million.

    \item We conduct an extensive empirical analysis where we examine the roles of (i) traditional / offline vs online incremental learning; (ii) ``look-back'' window size; (iii) feature extraction; (iv) memory size; (v) learning (neural network) vs statistical (ARIMA) models. Furthermore, all these are considered in three tasks (next-, 7-, and 14-day forecasting) and we provide a per-wave analysis.
\end{itemize}

The remainder of the paper is structured as follows. Section~\ref{sec:related} discusses work related to ours. Section~\ref{sec:method} describes the problem formulation and the incremental learning framework for adaptive forecasting. The experimental setup and results are provided in Section \ref{sec:exp_setup} and \ref{sec:exp_results} respectively. We conclude in Section~\ref{sec:conclusion}.

\section{Related Work}\label{sec:related}

\subsection{Compartmental models}
These models, like the well-known Susceptible Exposed Infectious Recovered (SEIR) \cite{doi:10.1098/rspa.1927.0118} and any variations of it, split the population into mutually exclusive states that describe a path of infection dynamics through mathematical modelling \cite{Li2022-bu}. For maximum accuracy, studies \cite{CALAFIORE2020361}, \cite{IHME_COVID-19_Forecasting_Team2021-il} have deduced parameters describing the transition between states that are time-varying capturing the social changes, medical advancements and non-pharmaceutical interventions during a pandemic \cite{IHME_COVID-19_Forecasting_Team2021-il}. For example, the "DELPHI" model \cite{Li2022-bu} consists of 11 compartments and forecasts detected cases and deaths for about 2 weeks, accounting for government measures and limited population testing. The vast majority of existing work on COVID-19 cases forecasting lie within this domain.

\subsection{Data-driven methods}
The focus of our work is on data-driven methods. Forecasting using data-driven methods can also be successfully achieved through statistical and machine learning methods. Isaac B. et al. \cite{Boyd2022-rx} compared the performance of a model that combines Convolutional and Long Short Term Memory (LSTM) layers to that of a standard neural network, using a 14-day window of positive cases to predict those of the next seven days both at the regional and national level. Several studies have compared the performance of LSTM to that of other models including Recurrent Neural Networks (RNNs) \cite{ALASSAFI2022335}, Gradient Boosting Trees \cite{Luo2021-ap} and the statistical model ARIMA \cite{Ketu2022-vt}. Research includes time series of just confirmed cases for generalizability \cite{Boyd2022-rx}, added features like number of cured patients and deaths \cite{Ketu2022-vt} and aggregated features \cite{Ketu2022-vt}, \cite{Luo2021-ap} for improved accuracy. The superiority of the LSTM is concluded in all the last comparisons. Another comparative study in \cite{ZEROUAL2020110121} used LSTM, RNN, Bidirectional LSTM, Gated Recurrent Units (GRUs) and Variational AutoEncoder (VAE) to predict new and recovered cases for the next 17-days where VAE showed the best performance. Encoders of self-attention and recurrent layers, that consider among other factors travelling from each country to predict the spread were also proposed in \cite{10.1145/3394486.3412864}.

The aforementioned methods consider offline learning. Continual or online learning is starting to be used to capture the concept of drift in the spread of COVID-19 and adapt models in real time. The study in \cite{UCHIDA2022380} evaluates the best number of training samples needed at each time step to minimize the prediction error and, thus, capture drift. Ridge regression is used for predictions of hospitalizations from new cases, severe cases from hospitalizations and deaths from severe cases for the next 7 days using 14-day windows \cite{UCHIDA2022380}. In \cite{9868791} an ensemble of regression models predicts 30-day mortality allowing for adaptation of the models at every instance by i) fitting them again on the whole data, ii) fitting them again on just the new instance, or iii) fitting a completely new ensemble on the new data. A linear model with LASSO  (least absolute shrinkage and selection operator) penalty \cite{Liu2020-hu} and a feed-forward network with autoregressive input (predictions at each time step used for training for the next forecast) \cite{Rodríguez_Tabassum_Cui_Xie_Ho_Agarwal_Adhikari_Prakash_2021}, were also able to incrementally train and produce 2-day cases predictions  \cite{Liu2020-hu} and 30-day predictions of hospitalizations and deaths \cite{Rodríguez_Tabassum_Cui_Xie_Ho_Agarwal_Adhikari_Prakash_2021}, respectively.

\subsection{Hybrid}
A study \cite{Camargo2022-kq} has used data from a compartmental model (exposed, infected, recovered and dead population) to evaluate the best lags of each out of time series windows in an ARIMA model, for predicting each of the variables and susceptible population. Based on this, new data is then continuously bootstrapped out of a data stream, predicting and updating incrementally an ensemble of algorithms each time \cite{Camargo2022-kq}. In \cite{Farooq2020-bi} and \cite{Farooq2021-hq} incremental learning of a neural network provides 5 parameters (rate of infection during lockdown, time lockdown begins, rate of death, rate of recovery) needed for a Susceptible Infected Recovered Vaccinated Deceased (SIRVD) model. The SIRVD model forecasts monthly trajectories of deaths under different senarios \cite{Farooq2020-bi} and monthly total number of cases, active infections and deaths \cite{Farooq2021-hq}.

\section{Incremental Learning Framework for Adaptive Forecasting}\label{sec:method}
We consider a data generating process $S = \{n^t\}_{t=1}^{T}$ that provides at each day $t$ a number $n^t \in \mathbb{R}$ of positive cases, from an unknown and evolving probability distribution $p^t(n)$, where $T \in [1,\infty)$. The instances constitute a univariate time series, and $n^t$ corresponds to the number of COVID-19 cases on day $t$.

To address the temporal aspects of the data, we consider a sliding window of size $W \in \mathbb{Z}^+$, such that, $x^t = \{n^t, n^{t-1}, ..., n^{t-W+1}\} \in \mathbb{R}^W$ is a $W$-dimensional vector belonging to input space $X \subset \mathbb{R}^W$. The task is to forecast $D \in \mathbb{Z}^+$ days of the COVID-19 cases, that is, at any day $t > W$ to predict $\hat{y}^{t+D} = \{\hat{n}^{t+D}, ..., \hat{n}^{t+2}, \hat{n}^{t+1}\} \in \mathbb{R}^D$, a $D$-dimensional vector belonging to $Y \subset \mathbb{R}^D$.

\begin{algorithm}[t!]
	\caption{Data-driven framework for adaptive forecasting}
	\label{alg:method}
	\begin{algorithmic}[1]
            \Statex \textbf{Input:}
            \Statex $D$: number of days to forecast
            \Statex $W$: ``Lookback'' window size
            \Statex $M$: Memory / queue size

            \State Wait $W$ days, observe instance $x^W = \{n^1, ..., n^W\}$
            \State Create model $f^W.init()$
            \State Predict $\hat{y}^{W+1} = f^W.predict(x^W)$
            \For{each time step $t \in [W + 1, W + D - 1]$}
                \State Get ground truth $y^t = n^t$

                \State Observe instance $x^t = \{n^t, n^{t-1}, .., n^{t-W+1}\}$
                \State Predict $\hat{y}^{t+1} = f^W.predict(x^t)$

            \EndFor
                \State Initialise memory $q^t = \{\}$
    		\For{each time step $t \in [W + D, \infty)$}
                \State Get ground truth $y^t = n^t$
                \State Observe $y^t = \{n^t, n^{t-1}, .., n^{t-D+1}\}$
                \State Append example to memory $q^t = q^{t-1}.append((x^{t-D}, y^t))$
                \State Incremental training $f^t = f^{t-1}.train(q^t)$
                \State Observe instance $x^t = \{n^t, n^{t-1}, .., n^{t-W+1}\}$
                \State Predict $\hat{y}^{t + 1} = f^t.predict(x^t)$
		    \EndFor
		
	\end{algorithmic}
\end{algorithm}

A regression model $f^t$ receives a new example $x^t \in \mathbb{R}^W$ at time step $t$ and makes a prediction $\hat{y}^{t+D} \in \mathbb{R}^D$, based on a concept $f: X \to Y$ such that $\hat{y}^{t+D} = f^t(x^t)$. In this study, we will be using neural networks as our regression models, which they have been demonstrated to be effective incremental learners \cite{losing2018incremental} \cite{malialis2020online} \cite{malialis2022nonstationary}. The loss function used between a prediction $\hat{y}^t \in \mathbb{R}^D$ and ground truth $y^t \in \mathbb{R}^D$ at time $t$ is the Mean Squared Error (MSE) defined as:
\begin{equation}\label{eq:mae}
    J^t = MSE(\hat{y}^t, y^t) = \frac{1}{D} \sum_{d=1}^{D} (\hat{y}^t_d - y^t_d)^2,
\end{equation}

The model is continually updated using incremental learning, which is defined as the gradual adaptation of a model without complete re-training, that is, $f^{t} = f^{t-1}.train((x^{t-D}, y^{t}))$. Learning is performed using incremental Stochastic Gradient Descent where each neural network weight $w$
is updated according to the formula $w^t \leftarrow w^{t-1} - \alpha \frac{\partial{J^t}}{w}$, where $\frac{\partial{J^t}}{w}$ is the partial derivative with respect to $w$, and $\alpha$ is the learning rate.

Furthermore, we introduce a memory component implemented as a queue $q$ of size $M$, which stores historical examples. For instance, at time $t$, it will append to memory the example $(x^{t-D}, y^{t})$, i.e., $m^{t} = m^{t-1}.append((x^{t-D}, y^{t}))$. As a result, incremental learning is now performed using $f^{t} = f^{t-1}.train(q^t)$, and the loss function is defined as the average MSE of all memory examples.

The framework's pseudocode is shown in Alg.~\ref{alg:method}. Initially, we wait for $W$ days (Line 1). Subsequently and until day $t < W + D$, we only perform prediction (i.e., forecasting) without any incremental training (Lines 2 - 7). From day $t \geq W + D$ we perform both prediction and incremental learning (Lines 8 - 15).

\section{Experimental Setup}\label{sec:exp_setup}

\subsection{Dataset}\label{sec:exp_datasets}

\begin{figure}[t]
    \centering
    \includegraphics[width=0.7\textwidth]{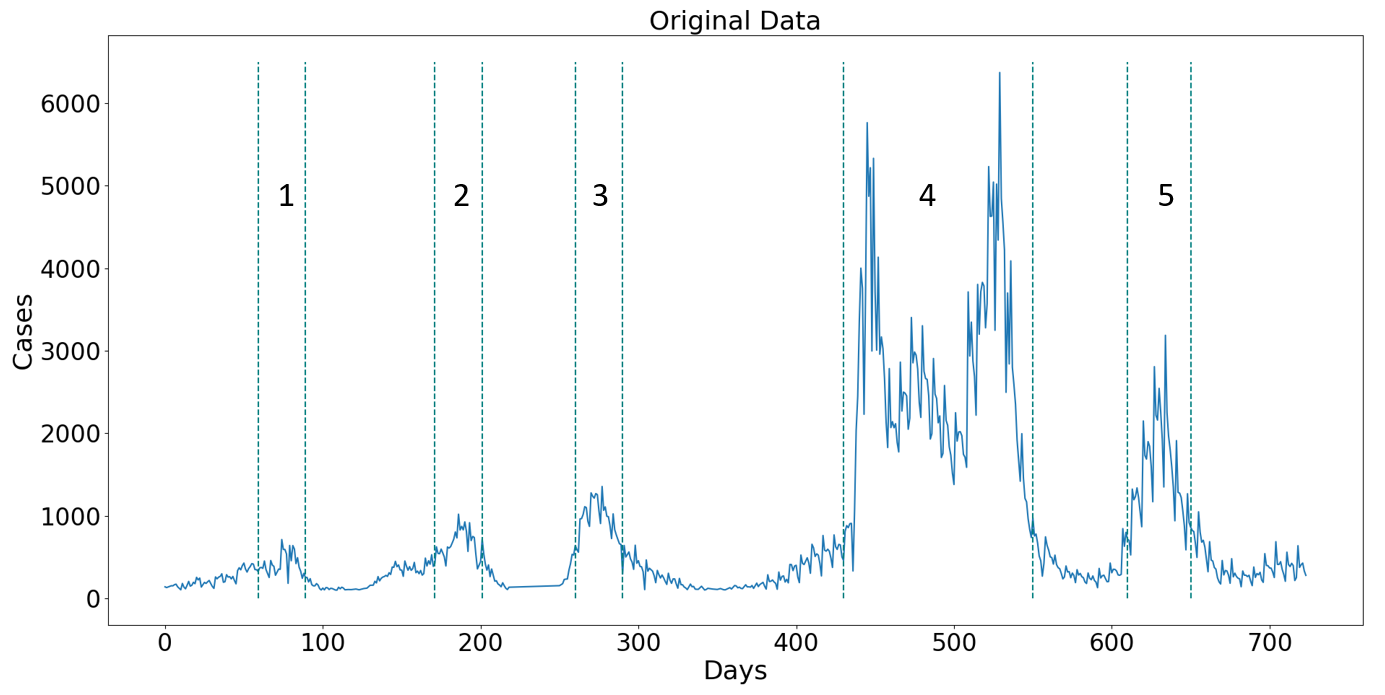}
    \caption{COVID-19 cases in Cyprus (15/10/20-18/10/22)}
    \label{fig:rawcases}
\end{figure}

Our data consist of reported daily SARS-CoV-2 cases in Cyprus from 15/10/20 to 08/10/22 as they appear in the TESSy platform of the European Center of Disease Prevention and Control (ECDC) in the RESPISURV dataset. Data preprocessing included creating sliding windows of 7, 14, 30 days and removal of daily cases of under 100 for reduced noise and easier learning of the models. Any missing values were imputed with the mean of their row or the previous row and data were normalized by dividing by maximum number of cases. Six periods of interest, referred to as ``waves'', are considered in this study and are shown in Figure \ref{fig:rawcases}. The time periods of each wave are as follows: Wave 1: 13/12/20 - 11/01/21; Wave 2: 04/04/21 - 03/05/21; Wave 3: 02/07/21 - 31/07/21; Wave 4: 19/12/21 - 07/01/22; Wave 5: 17/06/22 - 26/07/22. Also, we will be referring to the remaining (i.e., non-waves) time period as ``normal''.

\subsection{Compared methods}\label{sec:exp_methods}
The compared methods follow the same framework shown in Algorithm~\ref{alg:method}.

\textbf{MLP}. It refers to the standard feed-forward, fully-connected Multilayer Perceptron (MLP) model. In all experiments, its hyper-parameters are: He Normal \cite{he2015delving} weight initialisation, the Adam \cite{kingma2014adam} optimisation algorithm, LeakyReLU \cite{maas2013rectifier} and ReLU for the hidden and output activation function respectively, the MSE loss function, and mini-batch size of one. The rest of them (architecture, learning rate, regularisation, and number of epochs) slightly vary for each experiment.

\textbf{ARIMA}. The Autoregressive Integrated Moving Average (ARIMA) model is a classical forecasting method. Despite the fact that ARIMA is often considered as a baseline method, it is emphasised that due to the limited availability of historical data, it is actually demonstrated to be very effective particularly during normal and small outbreaks. In all experiments, its hyper-parameters are number of lagged observations for auto-regression: 1, number of times the raw observations are differenced: 0 and moving average window size: 0.

\subsection{Evaluation method and metrics}\label{sec:eval}
To evaluate and compare the aforementioned methods, we have been using the following widely adopted metrics for regression forecasts.

\textbf{MAE}. This refers to the Mean Absolute Error (MAE).

\textbf{MAPE}. This refers to the Mean Absolute Percentage Error (MAPE) between actual $y^t \in \mathbb{R}^D$ and predicted $\hat{y}^t \in \mathbb{R}^D$ values as defined below:
\begin{equation}
MAPE(\hat{y}, y) = \frac{100\%}{N} \sum_{d=1}^{D} \frac{|y_d - \hat{y}_d|}{|y_d|},   
\end{equation}

For all experiments involving neural networks, we run each one over 10 repetitions and provide the average and standard deviation for both metrics, for an overall time period, as well as during wave and normal periods.

\section{Experimental Results}\label{sec:exp_results}

\begin{figure}[t]
    \centering
    \includegraphics[width=0.7\textwidth]{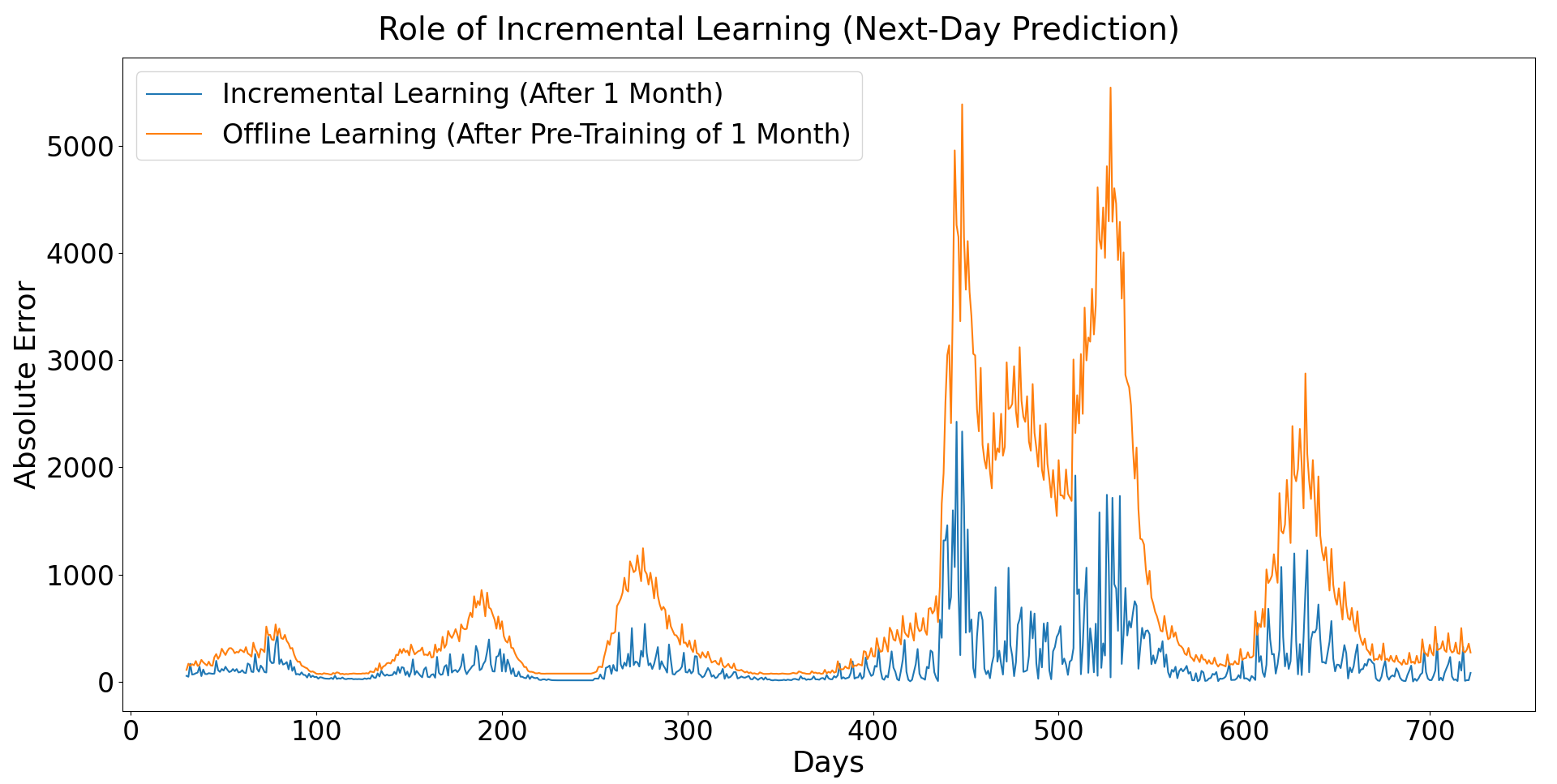}
    \caption{Online vs offline learning for next-day predictions}
    \label{fig:onlinevsofflinepred1}
\end{figure}

\subsection{Role of incremental learning}
This section compares the performance of MLP using offline learning to that of online incremental learning. For offline learning, the MLP was pre-trained on one month of data and with no further training. For the next-day prediction task, Figure~\ref{fig:onlinevsofflinepred1} shows the results of the two paradigms on a daily basis, while Table~\ref{tab:onlinevsofflinepred1} provides the relevant aggregated metrics. The standard deviation of the error is shown in brackets. The corresponding results for the 7- and 14-day forecasting tasks are shown in Table~\ref{tab:onlinevsofflinepred7} and Table \ref{tab:onlinevsofflinepred14}. It is observed that online learning significantly outperforms offline learning in all tasks and all time periods.

\begin{table}[t!]
\caption{MLP with online vs offline learning for next-day predictions}
\label{tab:onlinevsofflinepred1}
\begin{tabular}{|c|cc|cc|cc|}
\hline
\textbf{} & \multicolumn{2}{c|}{\textbf{Overall}} & \multicolumn{2}{c|}{\textbf{Waves}} & \multicolumn{2}{c|}{\textbf{Normal}} \\ \hline
\textbf{} & \multicolumn{1}{c|}{\textbf{MAE}} & \textbf{MAPE} & \multicolumn{1}{c|}{\textbf{MAE}} & \textbf{MAPE} & \multicolumn{1}{c|}{\textbf{MAE}} & \textbf{MAPE} \\ \hline
\textbf{Online} & \multicolumn{1}{c|}{\textbf{186.1 (47.4)}} & \textbf{26.3 (14.3)} & \multicolumn{1}{c|}{\textbf{378.4 (62.5)}} & \textbf{25.5 (9.1)} & \multicolumn{1}{c|}{\textbf{148.2 (50.4)}} & \textbf{26.7 (17.9)} \\ \hline
\textbf{Offline} & \multicolumn{1}{c|}{767.9 (405.3)} & 83.9 (43.8) & \multicolumn{1}{c|}{1714.1 (934.0)} & 96.7 (57.1) & \multicolumn{1}{c|}{588.2 (302.7)} & 79.9 (40.5) \\ \hline
\end{tabular}
\end{table}

\begin{table}[h!]
\caption{MLP with online vs offline learning for 7-day predictions}
\label{tab:onlinevsofflinepred7}
\begin{tabular}{|c|cc|cc|cc|}
\hline
\textbf{} & \multicolumn{2}{c|}{\textbf{Overall}} & \multicolumn{2}{c|}{\textbf{Waves}} & \multicolumn{2}{c|}{\textbf{Normal}} \\ \hline
\textbf{} & \multicolumn{1}{c|}{\textbf{MAE}} & \textbf{MAPE} & \multicolumn{1}{c|}{\textbf{MAE}} & \textbf{MAPE} & \multicolumn{1}{c|}{\textbf{MAE}} & \textbf{MAPE} \\ \hline
\textbf{Online} & \multicolumn{1}{c|}{\textbf{342.3 (48.0)}} & \textbf{43.7 (5.3)} & \multicolumn{1}{c|}{\textbf{714.1 (105.8)}} & \textbf{46.0 (5.0)} & \multicolumn{1}{c|}{\textbf{267.9 (36.5)}} & \textbf{43.5 (5.5)} \\ \hline
\textbf{Offline} & \multicolumn{1}{c|}{598.7 (59.4)} & 65.4 (6.3) & \multicolumn{1}{c|}{1319.8 (140.0)} & 75.1 (7.3) & \multicolumn{1}{c|}{461.2 (45.0)} & 62.7 (6.5) \\ \hline
\end{tabular}
\end{table}

\begin{table}[h!]
\caption{MLP with online vs offline learning for 14-day predictions}
\label{tab:onlinevsofflinepred14}
\begin{tabular}{|c|cc|cc|cc|}
\hline
 & \multicolumn{2}{c|}{\textbf{Overall}} & \multicolumn{2}{c|}{\textbf{Waves}} & \multicolumn{2}{c|}{\textbf{Normal}} \\ \hline
\textbf{} & \multicolumn{1}{c|}{\textbf{MAE}} & \textbf{MAPE} & \multicolumn{1}{c|}{\textbf{MAE}} & \textbf{MAPE} & \multicolumn{1}{c|}{\textbf{MAE}} & \textbf{MAPE} \\ \hline
\textbf{Online} & \multicolumn{1}{c|}{\textbf{524.2 (42.1)}} & \textbf{75.9 (4.1)} & \multicolumn{1}{c|}{\textbf{1011.1 (97.9)}} & \textbf{67.9 (4.9)} & \multicolumn{1}{c|}{\textbf{413.6 (31.6)}} & \textbf{74.5 (4.3)} \\ \hline
\textbf{Offline} & \multicolumn{1}{c|}{686.7 (47.2)} & 83.6 (7.8) & \multicolumn{1}{c|}{1416.5 (101.4)} & 88.3 (10.4) & \multicolumn{1}{c|}{546.3 (35.5)} & 82.0 (6.9) \\ \hline
\end{tabular}
\end{table}

\begin{table}[t]
\caption{MLP with 7-, 14- and 30-day sliding window (next-day prediction)}
\label{tab:mlpwindowspred1}
\begin{tabular}{|c|cc|cc|cl|}
\hline
\textbf{} & \multicolumn{2}{c|}{\textbf{Overall}} & \multicolumn{2}{c|}{\textbf{Waves}} & \multicolumn{2}{c|}{\textbf{Normal}} \\ \hline
\textbf{Window} & \multicolumn{1}{c|}{\textbf{MAE}} & \textbf{MAPE} & \multicolumn{1}{c|}{\textbf{MAE}} & \textbf{MAPE} & \multicolumn{1}{c|}{\textbf{MAE}} & \textbf{MAPE} \\ \hline
\textbf{7} & \multicolumn{1}{c|}{\textbf{186.1 (47.4)}} & \textbf{26.3 (14.3)} & \multicolumn{1}{c|}{378.4 (62.5)} & 25.5 (9.1) & \multicolumn{1}{c|}{\textbf{148.2 (50.4)}} & \textbf{26.7 (17.9)} \\ \hline
\textbf{14} & \multicolumn{1}{c|}{343.9 (258.0)} & 48.2 (28.8) & \multicolumn{1}{c|}{698.6 (569.1)} & 42.5 (30.9) & \multicolumn{1}{c|}{270.0 (201.5)} & 47.4 (29.9) \\ \hline
\textbf{30} & \multicolumn{1}{l|}{187.8 (20.8)} & \multicolumn{1}{l|}{28.3 (7.1)} & \multicolumn{1}{l|}{\textbf{368.5 (23.0)}} & \multicolumn{1}{l|}{\textbf{25.2 (5.2)}} & \multicolumn{1}{l|}{154.1 (23.6)} & 28.6 (8.7) \\ \hline
\end{tabular}
\end{table}

\begin{table}[h!]
\caption{MLP with 7-, 14- and 30-day sliding window (7-day prediction)}
\label{tab:mlpwindowspred7}
\begin{tabular}{|c|cc|cc|cl|}
\hline
\textbf{} & \multicolumn{2}{c|}{\textbf{Overall}} & \multicolumn{2}{c|}{\textbf{Waves}} & \multicolumn{2}{c|}{\textbf{Normal}} \\ \hline
\textbf{Window} & \multicolumn{1}{c|}{\textbf{MAE}} & \textbf{MAPE} & \multicolumn{1}{c|}{\textbf{MAE}} & \textbf{MAPE} & \multicolumn{1}{c|}{\textbf{MAE}} & \textbf{MAPE} \\ \hline
\textbf{7} & \multicolumn{1}{c|}{\textbf{342.3 (48.0)}} & \textbf{43.7 (5.3)} & \multicolumn{1}{c|}{714.1 (105.8)} & 46.0 (5.0) & \multicolumn{1}{c|}{\textbf{267.9 (36.5)}} & \textbf{43.5 (5.5)} \\ \hline
\textbf{14} & \multicolumn{1}{c|}{378.0 (43.5)} & 46.5 (4.7) & \multicolumn{1}{c|}{795.5 (95.5)} & 48.8 (4.9) & \multicolumn{1}{c|}{292.6 (32.9)} & 45.2 (4.6) \\ \hline
\textbf{30} & \multicolumn{1}{l|}{350.6 (38.9)} & \multicolumn{1}{l|}{46.4 (5.0)} & \multicolumn{1}{l|}{\textbf{709.2 (81.1)}} & \multicolumn{1}{l|}{\textbf{44.9 (4.7)}} & \multicolumn{1}{l|}{281.6 (32.1)} & 45.9 (5.5) \\ \hline
\end{tabular}
\end{table}

\subsection{Role of the sliding window size}
This section examines the impact of the sliding window size on the performance of MLP using incremental learning. Results are provided in Table \ref{tab:mlpwindowspred1} and Table \ref{tab:mlpwindowspred7} for next- and seven-day prediction tasks, respectively.

Using MLP, the 7-day window performs better compared to the 14-day and 30-day ones for all prediction tasks. Regarding average performance across waves, the 30-day window performs best for next-day and 7-day prediction and the 7-day window performs best for 14-day prediction task (not shown here). Normal periods benefit the most from a 7-day window for all prediction tasks. While not shown due to space restrictions, for ARIMA, a 30-day window performs the best on the overall data, wave and normal periods for all tasks.

The better performance of a 7-day window can be attributed to the fewer window days suggesting more recent data, which can increase performance. On the other hand, it is speculated that a 30-day window works best because of the more data and fluctuations considered.

\subsection{Role of feature extraction}\label{sec:features}
This section describes the role of 20 features in our model, aggregated across a 14-day window. The features are: school closing strictness (mean), public events cancellation strictness (mean), positive cases (min, max), unvaccinated cases (min, median), second dose vaccinated population (min, range), second dose vaccinated cases (mean, median), first dose vaccinated cases (median, mean), weekly deaths (mean), workplace closing strictness (mean), weekly ICU cases (mean), weighted stringency index (median), recovered (s.d.), 70+ aged cases (mean), first dose vaccinated population (median) and 18-24 aged cases (mean).

The results for next- and seven-day prediction tasks are shown in Table \ref{tab:mlpfeaturespred1} and Table \ref{tab:mlpfeaturespred7}, respectively. Using the features, MAPE is reduced by 8.3\%, 7.2\% and 0.2\% for overall, wave and normal periods, respectively, for the 7-day prediction. The features seem to be more informative when making later predictions.

\begin{table}[t]
\caption{MLP with raw data vs features (next-day prediction)}
\label{tab:mlpfeaturespred1}
\begin{tabular}{|c|cc|cc|cl|}
\hline
\textbf{} & \multicolumn{2}{c|}{\textbf{Overall}} & \multicolumn{2}{c|}{\textbf{Waves}} & \multicolumn{2}{c|}{\textbf{Normal}} \\ \hline
\textbf{Data} & \multicolumn{1}{c|}{\textbf{MAE}} & \textbf{MAPE} & \multicolumn{1}{c|}{\textbf{MAE}} & \textbf{MAPE} & \multicolumn{1}{c|}{\textbf{MAE}} & \textbf{MAPE} \\ \hline
\textbf{Raw} & \multicolumn{1}{c|}{\textbf{186.1 (47.4)}} & \textbf{26.3 (14.3)} & \multicolumn{1}{c|}{\textbf{378.4 (62.5)}} & \textbf{25.5 (9.1)} & \multicolumn{1}{c|}{\textbf{148.2 (50.4)}} & \textbf{26.7 (17.9)} \\ \hline
\textbf{Features} & \multicolumn{1}{c|}{211.5 (42.1)} & 30.5 (11.8) & \multicolumn{1}{c|}{423.2 (57.1)} & 29.4 (9.1) & \multicolumn{1}{c|}{172.0 (43.8)} & 31.8 (14.1) \\ \hline
\end{tabular}
\end{table}

\begin{table}[h!]
\caption{MLP with raw data vs features (7-day prediction)}
\label{tab:mlpfeaturespred7}
\begin{tabular}{|c|cc|cc|cl|}
\hline
\textbf{} & \multicolumn{2}{c|}{\textbf{Overall}} & \multicolumn{2}{c|}{\textbf{Waves}} & \multicolumn{2}{c|}{\textbf{Normal}} \\ \hline
\textbf{Data} & \multicolumn{1}{c|}{\textbf{MAE}} & \textbf{MAPE} & \multicolumn{1}{c|}{\textbf{MAE}} & \textbf{MAPE} & \multicolumn{1}{c|}{\textbf{MAE}} & \textbf{MAPE} \\ \hline
\textbf{Raw} & \multicolumn{1}{c|}{342.3 (48.0)} & 43.7 (5.3) & \multicolumn{1}{c|}{714.1 (105.8)} & 46.0 (5.0) & \multicolumn{1}{c|}{267.9 (36.5)} & 43.5 (5.5) \\ \hline
\textbf{Features} & \multicolumn{1}{c|}{\textbf{284.7 (11.1)}} & \textbf{35.4 (2.9)} & \multicolumn{1}{c|}{\textbf{603.1 (15.4)}} & \textbf{38.8 (2.0)} & \multicolumn{1}{c|}{\textbf{238.7 (13.5)}} & \textbf{43.3 (5.5)} \\ \hline
\end{tabular}
\end{table}

\subsection{Role of the memory size}
In this section, the role of the memory size using i) raw data and ii) features is assessed. For these experiments, raw data was used in 7-day windows for 14-day predictions and features were extracted from 14-day windows for 7-day predictions. Window size here is chosen based on best windows for raw data and features, respectively, as stated in Sections 5.2 and 5.3. The results for the memory use with raw data are shown in Table \ref{tab:mlpmemorywin7pred14} and with features in Table \ref{tab:mlpmemoryfeatswin14pred7}.

In the first case, it is deduced that increasing memory size improves overall, wave and normal periods performance by up to 17\%, 12.7\% and 16.1\%, respectively. Interestingly, in the second case, using memory decreases performance.

\begin{table}[t!]
\caption{MLP performance with raw data per memory size (7-day window, 14-day prediction)}
\label{tab:mlpmemorywin7pred14}
\begin{tabular}{|c|cc|cc|cc|}
\hline
\textbf{} & \multicolumn{2}{c|}{\textbf{Overall}} & \multicolumn{2}{c|}{\textbf{Waves}} & \multicolumn{2}{c|}{\textbf{Normal}} \\ \hline
\textbf{Memory} & \multicolumn{1}{c|}{\textbf{MAE}} & \textbf{MAPE} & \multicolumn{1}{c|}{\textbf{MAE}} & \textbf{MAPE} & \multicolumn{1}{c|}{\textbf{MAE}} & \textbf{MAPE} \\ \hline
\textbf{1} & \multicolumn{1}{c|}{524.2 (42.1)} & 75.9 (4.1) & \multicolumn{1}{c|}{1011.1 (97.9)} & 67.9 (4.9) & \multicolumn{1}{c|}{413.6 (31.6)} & 74.5 (4.3) \\ \hline
\textbf{30} & \multicolumn{1}{c|}{570.0 (35.7)} & 98.7 (3.8) & \multicolumn{1}{c|}{981.1 (93.9)} & 67.9 (4.1) & \multicolumn{1}{c|}{464.3 (27.5)} & 98.2 (4.4) \\ \hline
\textbf{90} & \multicolumn{1}{c|}{505.9 (50.7)} & 78.9 (5.8) & \multicolumn{1}{c|}{917.8 (106.9)} & 63.4 (5.0) & \multicolumn{1}{c|}{421.8 (38.9)} & 82.9 (6.2) \\ \hline
\textbf{180} & \multicolumn{1}{c|}{428.7 (35.2)} & 65.5 (3.4) & \multicolumn{1}{c|}{798.2 (77.2)} & 57.8 (3.9) & \multicolumn{1}{c|}{350.3 (25.5)} & 67.0 (3.2) \\ \hline
\textbf{240} & \multicolumn{1}{c|}{445.1 (40.4)} & 65.6 (3.7) & \multicolumn{1}{c|}{846.8 (90.6)} & 59.0 (4.3) & \multicolumn{1}{c|}{349.8 (32.1)} & 63.9 (4.0) \\ \hline
\textbf{360} & \multicolumn{1}{c|}{\textbf{418.7 (46.1)}} & \textbf{58.9 (4.3)} & \multicolumn{1}{c|}{\textbf{820.7 (97.7)}} & \textbf{55.2 (5.2)} & \multicolumn{1}{c|}{\textbf{329.3 (36.6)}} & \textbf{58.4 (4.7)} \\ \hline
\end{tabular}
\end{table}

\begin{table}[h!]
\caption{MLP performance with features per memory size (14-day window, 7-day prediction)}
\label{tab:mlpmemoryfeatswin14pred7}
\begin{tabular}{|c|cc|cc|cc|}
\hline
\textbf{} & \multicolumn{2}{c|}{\textbf{Overall}} & \multicolumn{2}{c|}{\textbf{Waves}} & \multicolumn{2}{c|}{\textbf{Normal}} \\ \hline
\textbf{Memory} & \multicolumn{1}{c|}{\textbf{MAE}} & \textbf{MAPE} & \multicolumn{1}{c|}{\textbf{MAE}} & \textbf{MAPE} & \multicolumn{1}{c|}{\textbf{MAE}} & \textbf{MAPE} \\ \hline
\textbf{1} & \multicolumn{1}{c|}{\textbf{284.7 (11.1)}} & \textbf{35.4 (2.9)} & \multicolumn{1}{c|}{\textbf{603.1 (15.4)}} & \textbf{38.8 (2.0)} & \multicolumn{1}{c|}{\textbf{238.7 (13.5)}} & \textbf{43.3 (5.5)} \\ \hline
\textbf{30} & \multicolumn{1}{c|}{513.8 (13.4)} & 84.9 (0.9) & \multicolumn{1}{c|}{880.1 (37.3)} & 56.1 (2.0) & \multicolumn{1}{c|}{440.2 (11.3)} & 94.4 (4.4) \\ \hline
\textbf{90} & \multicolumn{1}{c|}{622.8 (3.7)} & 132.2 (0.9) & \multicolumn{1}{c|}{912.2 (5.6)} & 51.9 (1.4) & \multicolumn{1}{c|}{581.1 (13.6)} & 158.2 (7.1) \\ \hline
\textbf{180} & \multicolumn{1}{c|}{702.7 (2.0)} & 142.5 (0.6) & \multicolumn{1}{c|}{1071.6 (2.2)} & 57.2 (0.6) & \multicolumn{1}{c|}{580.3 (7.9)} & 147.6 (4.5) \\ \hline
\textbf{240} & \multicolumn{1}{c|}{696.7 (3.5)} & 134.2 (0.9) & \multicolumn{1}{c|}{1115.2 (5.5)} & 57.3 (1.4) & \multicolumn{1}{c|}{537.8 (13.8)} & 130.1 (6.9) \\ \hline
\textbf{360} & \multicolumn{1}{c|}{651.0 (2.2)} & 110.4 (0.6) & \multicolumn{1}{c|}{1181.4 (2.5)} & 57.0 (0.6) & \multicolumn{1}{c|}{496.7 (8.1)} & 107.7 (4.6) \\ \hline
\end{tabular}
\end{table}

\subsection{Comparative study}
\begin{figure}[t!]
    \centering
    \includegraphics[width=0.7\textwidth]{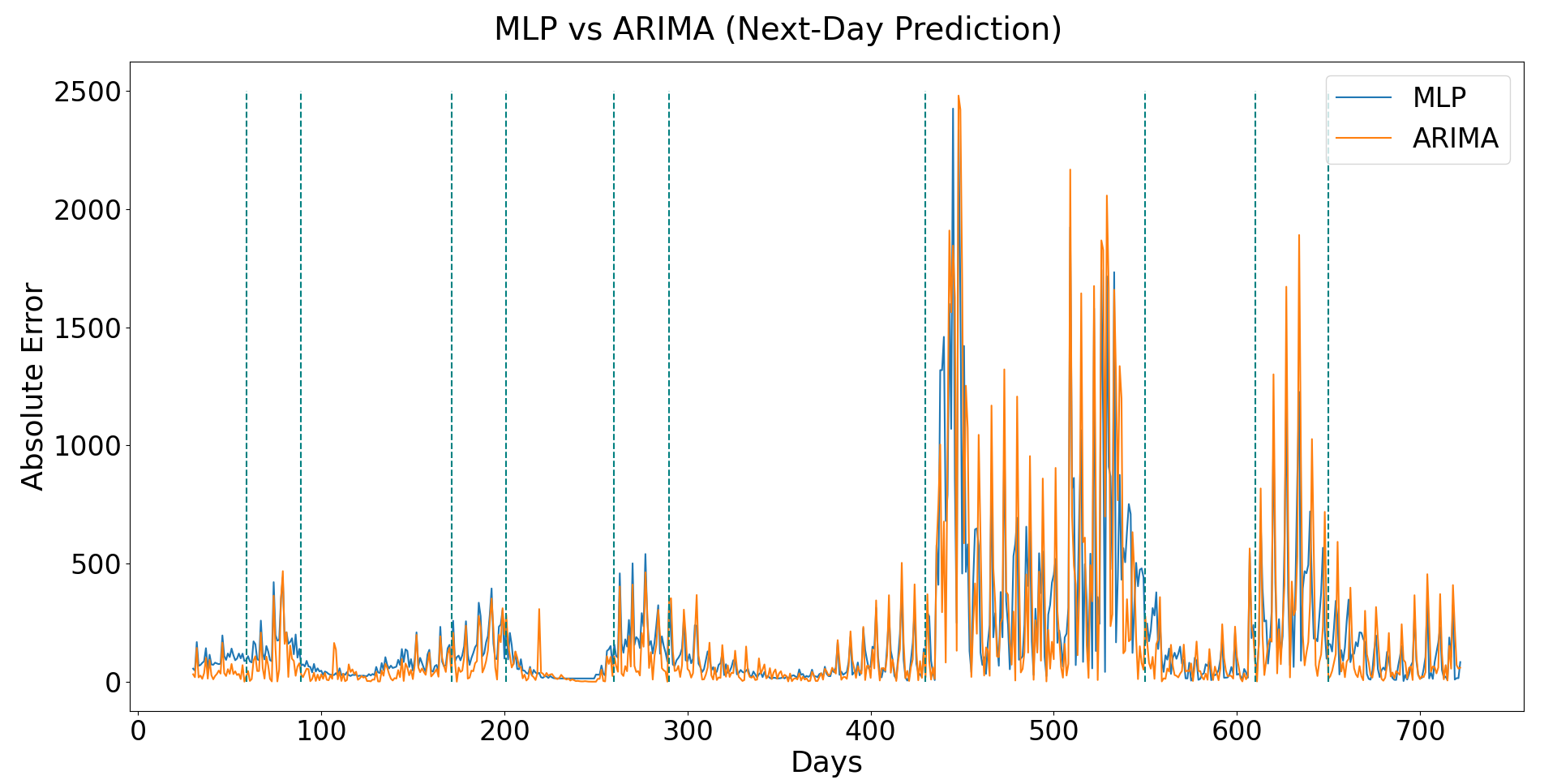}
    \caption{MLP vs ARIMA (next-day prediction)}
    \label{fig:mlparimapred1}
\end{figure}

This section aims to compare the best MLP experiments in this study to the traditional forecasting ARIMA method. Results refer to overall, wave and normal periods, as well as each wave. For next-day predictions, they are reported in Table \ref{tab:mlparimapred1} and Table \ref{tab:mlparimaperwavepred1}, and for 14-day predictions in Table \ref{tab:mlparimapred14} and Table \ref{tab:mlparimaperwavepred14}. Next-day prediction learning curves for the two models are shown in Figure \ref{fig:mlparimapred1}.
 
It is observed that for next-day predictions, the neural network outperforms ARIMA at Waves 4 and Wave 5 (Table \ref{tab:mlparimaperwavepred1}), with MAE of 529.1 (against 557.1) at Wave 4 and MAE of 371.5 (against 385.4) at Wave 5. For 14-day predictions, MLP captures the data distribution shift at Wave 4 (Table \ref{tab:mlparimaperwavepred14}) better than ARIMA with MAE of 1209 (against 1433.5).

\section{Conclusions and Future Work}\label{sec:conclusion}
The COVID-19 virus has been acutely affecting millions of people for more than three years. In valuable attempts for prompt government interventions and addressing data non-stationarity and availability, we have conducted an empirical study of data-driven (learning, statistical) methods using incremental training for adaptive forecasting of COVID-19 cases.
Some future directions are:

\textbf{Role of the memory}. The impact of memory is unclear. We have demonstrated its effectiveness on the performance of MLP with raw data, however, performance declined for MLP with features. Future work will investigate this.

\textbf{Statistical models with features}. In this study we examine the impact of features in MLP. We plan to use ARIMAX \cite{aji2021forecasting} to incorporate features to ARIMA.

\textbf{Advanced neural architectures}. Future work will investigate more complex neural architectures, such as, autoregressive networks and LSTMs.

\begin{table}[b!]
\caption{MLP vs ARIMA (next-day prediction)}
\label{tab:mlparimapred1}
\begin{tabular}{|c|cc|cc|cc|}
\hline
\textbf{} & \multicolumn{2}{c|}{\textbf{Overall}} & \multicolumn{2}{c|}{\textbf{Waves}} & \multicolumn{2}{c|}{\textbf{Normal}} \\ \hline
\textbf{Model} & \multicolumn{1}{c|}{\textbf{MAE}} & \textbf{MAPE} & \multicolumn{1}{c|}{\textbf{MAE}} & \textbf{MAPE} & \multicolumn{1}{c|}{\textbf{MAE}} & \textbf{MAPE} \\ \hline
\textbf{MLP} & \multicolumn{1}{c|}{186.1 (47.4)} & 26.3 (14.3) & \multicolumn{1}{c|}{378.4 (62.5)} & 25.5 (9.1) & \multicolumn{1}{c|}{148.2 (50.4)} & 26.7 (17.9) \\ \hline
\textbf{ARIMA} & \multicolumn{1}{c|}{\textbf{176.2}} & \textbf{20.3} & \multicolumn{1}{c|}{\textbf{371.7}} & \textbf{20.5} & \multicolumn{1}{c|}{\textbf{137.6}} & \textbf{19.6} \\ \hline
\end{tabular}
\end{table}

\begin{table}[b!]
\caption{MLP vs ARIMA per wave (next-day prediction)}
\label{tab:mlparimaperwavepred1}
\begin{tabular}{|c|c|c|c|c|c|}
\hline
\textbf{} & \textbf{Wave 1} & \textbf{Wave 2} & \textbf{Wave 3} & \textbf{Wave 4} & \textbf{Wave 5} \\ \hline
\textbf{Model} & \textbf{MAE} & \textbf{MAE} & \textbf{MAE} & \textbf{MAE} & \textbf{MAE} \\ \hline
\textbf{MLP} & 167.3 (131.5) & 168.2 (174.4) & 206.4 (264.4) & \textbf{529.1 (13.8)} & \textbf{371.5 (18.2)} \\ \hline
\textbf{ARIMA} & \textbf{103.5} & \textbf{123.5} & \textbf{128.4} & 557.1 & 385.4 \\ \hline
\end{tabular}
\end{table}

\begin{table}[t!]
\caption{MLP vs ARIMA (14-day prediction)}
\label{tab:mlparimapred14}
\begin{tabular}{|c|cc|cc|cc|}
\hline
\textbf{} & \multicolumn{2}{c|}{\textbf{Overall}} & \multicolumn{2}{c|}{\textbf{Waves}} & \multicolumn{2}{c|}{\textbf{Normal}} \\ \hline
\textbf{Model} & \multicolumn{1}{c|}{\textbf{MAE}} & \textbf{MAPE} & \multicolumn{1}{c|}{\textbf{MAE}} & \textbf{MAPE} & \multicolumn{1}{c|}{\textbf{MAE}} & \textbf{MAPE} \\ \hline
\textbf{MLP} & \multicolumn{1}{c|}{418.7 (46.1)} & 58.9 (4.3) & \multicolumn{1}{c|}{\textbf{820.7 (97.7)}} & 55.2 (5.2) & \multicolumn{1}{c|}{329.3 (36.6)} & 58.4 (4.7) \\ \hline
\textbf{ARIMA} & \multicolumn{1}{c|}{\textbf{412.7}} & \textbf{42.8} & \multicolumn{1}{c|}{861.6} & \textbf{48.3} & \multicolumn{1}{c|}{\textbf{322.6}} & \textbf{40.9} \\ \hline
\end{tabular}
\end{table}

\begin{table}[h!]
\caption{MLP vs ARIMA per wave (14-day prediction)}
\label{tab:mlparimaperwavepred14}
\begin{tabular}{|c|c|c|c|c|c|}
\hline
\textbf{} & \textbf{Wave 1} & \textbf{Wave 2} & \textbf{Wave 3} & \textbf{Wave 4} & \textbf{Wave 5} \\ \hline
\textbf{Model} & \textbf{MAE} & \textbf{MAE} & \textbf{MAE} & \textbf{MAE} & \textbf{MAE} \\ \hline
\textbf{MLP} & 247.6 (29.2) & 339.8 (50.5) & 392.5 (58.8) & \textbf{1209.0 (153.5)} & 767.4 (65.4) \\ \hline
\textbf{ARIMA} & \textbf{142.9} & \textbf{246.3} & \textbf{297.9} & 1433.5 & \textbf{569.3} \\ \hline
\end{tabular}
\end{table}

\bibliographystyle{splncs04}
\bibliography{samplepaper}

\end{document}